\documentclass[10pt,twocolumn,letterpaper]{article}

\usepackage{cvpr}      %
\makeatletter
\@namedef{ver@everyshi.sty}{}
\makeatother
\usepackage{graphicx}
\usepackage{amsmath}
\usepackage{amssymb}
\usepackage{booktabs}
\usepackage{multicol}
\usepackage[table]{xcolor}
\usepackage[absolute]{textpos}
\usepackage{appendix}
\usepackage{siunitx}

\definecolor{gold}{RGB}{255,215,0}
\definecolor{silver}{RGB}{192,192,192}
\definecolor{bronze}{RGB}{205,127,50}

\usepackage[pagebackref,breaklinks,colorlinks]{hyperref}

\usepackage[capitalize]{cleveref}
\crefname{section}{Sec.}{Secs.}
\Crefname{section}{Section}{Sections}
\Crefname{table}{Table}{Tables}
\crefname{table}{Tab.}{Tabs.}

\def\cvprPaperID{****} %
\def\confName{CVPR}
\def\confYear{2023}

\begin{document}

\title{Neural Assets: Volumetric Object Capture and Rendering\\for Interactive Environments}
\author{
Alja\v{z} Bo\v{z}i\v{c}, Denis Gladkov, Luke Doukakis and Christoph Lassner\\
Meta Reality Labs Research\\
{\tt\small \{aljaz, denisg, lukedoukakis, classner\}@meta.com}
}
\twocolumn[{
	\renewcommand\twocolumn[1][]{#1}%
	\maketitle
	\begin{center}
	    \vspace*{-0.2cm}
		\includegraphics[width=0.98\textwidth]{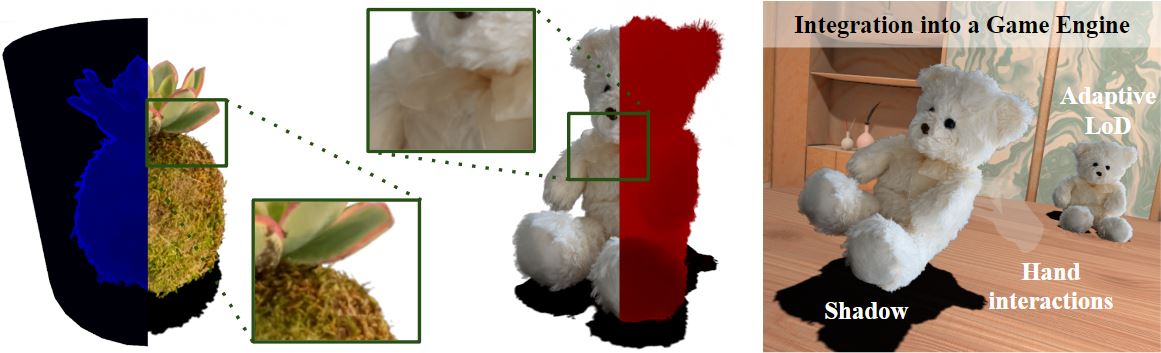}
		\vspace*{-0.2cm}
		\captionof{figure}{%
		We capture and render \textbf{Neural Assets}---real world objects scanned using a mobile device---volumetrically and seamlessly in interactive virtual environments.
		\textbf{Left/blue:} render volume and number-of-sample visualization.
		All pixels on the bounding volume are path traced to create the rendering, however for most pixels none (black) or few (dark blue) samples are needed to produce the rendering.
		Most samples are needed to render object boundaries (light blue).
		\textbf{Center/red:} objects with fur, flyaway hair and translucent parts, such as the bow tie, are represented correctly.
		The fur and bow tie blend seamlessly and realistically with the background.
		The details are reproduced in appearance \emph{and} geometry, which becomes apparent in the red depth map.
		\textbf{Right:} game engine integration and interaction.}
		\label{fig:teaser}
	\end{center}
}]

\pagestyle{plain}
\begin{textblock*}{1pt}(10cm, 25.7cm)1\end{textblock*}

\begin{abstract}
   Creating realistic virtual `assets' is a time-consuming process: it usually involves an artist designing the object, then spending a lot of effort on tweaking its appearance.
   Intricate details and certain effects, such as subsurface scattering, elude representation using real-time BRDFs, making it impossible to fully capture the appearance of certain objects.
   Inspired by the recent progress of neural rendering, we propose an approach for capturing real-world objects in everyday environments faithfully and fast.
   We use a novel neural representation to reconstruct volumetric effects, such as translucent object parts, and preserve photorealistic object appearance.
   To support real-time rendering without compromising rendering quality, our model uses a grid of features and a small MLP decoder that is transpiled into efficient shader code with interactive framerates.
   This leads to a seamless integration of the proposed neural assets with existing mesh environments and objects.
   Thanks to the use of standard shader code rendering is portable across many existing hardware and software systems.
   \vspace*{-0.3cm}
\end{abstract}

\section{Introduction}
\label{sec:introduction}
The appearance of virtual environments is only as convincing as their parts: the virtual `assets' that have been used to design them.
Hence, designers and artists spend a tremendous amount of time creating and fine-tuning object appearance and scene integration to match real scenes by finetuning object shape, as well as material properties and scene lighting.
This includes object scanning with expensive hardware, such as laser scanners, mesh `cleanup' (removal of scan artifacts), texture editing, BRDF definition, optimization and shader design.
The latter is particularly complex, because effects such as translucency and subsurface scattering are usually not (correctly) modeled in real-time rendering systems, leading to complications when the appearance of real-world objects should be reproduced.
In case of thin and organic materials such as skin or fur this results in `uncanny valley' effects present in many virtual productions.
In this paper, we introduce methods for reconstructing, rendering and integrating objects into interactive environments seamlessly using models that work with consumer hardware, have real-time performance, reproduce appearance faithfully and can be easily integrated with many game engine and graphics systems. 

We build on recent advances in neural volumetric rendering based on the seminal works on neural volumes~\cite{neural_volumes_Lombardi:2019} and radiance fields (NeRFs)~\cite{mildenhall2020nerf}.
Radiance fields are a volumetric representation that aims to directly model the plenoptic function of an object or scene, \ie, they directly model the emitted or reflected light (\ie, radiance) for every position in space depending on the view direction.
The recent success of neural representations is largely due to their elegant use as radiance field models that can be optimized end-to-end.
Overall, neural radiance fields have only minimal requirements for data capturing: a set of RGB images captured with a mobile phone is enough to to reconstruct a scene with high quality, as long as a decent camera calibration can be extracted.
Furthermore, they are versatile and powerful in modeling a wide variety of objects and scenes, including translucent materials and fur, that are hard or impossible to model using mesh-based representations.

However, our intended application is notably different from standard radiance field reconstruction: we aim to separate a foreground object from its background without simplifying assumptions (e.g., a green screen) and making the capture process as user-friendly as possible, ideally requiring only a ca.~90s of video capture.
Reconstructing fine details in the object boundary or translucency requires a full reconstruction of the scene.
To focus model capacity on the object of interest, we use two radiance field models: one for the background and one for the foreground, and only use the foreground model as asset.
To make the capture process as user-friendly as possible, we directly address many common recording mistakes in a simple capture application and fix, for example, white balance and exposure and let the user select the foreground object bounding box in 3D.

After reconstruction, neural volumetric models don't play well with existing interactive graphics systems: (1)~they use different rendering `primitives' that don't fit well into traditional graphics shading systems and (2)~they don't naturally plug in to the various game engine subsystems, such as the color blending, shadowing, level-of-detail (LoD) and physics systems.
We propose to overcome these problems by creating a transpiler that produces GLSL or HLSL shader code for a pixel shader.
The model architecture has been optimized to work particularly well in this setting and the shader is able to render the model in super real-time.
The shader performs volumetric raycasting, aggregates sample points and produces a result that blends correctly with the background consisting of other objects, mesh- or volume-based.
We fuse the asset with a marching-cubes generated, low-fidelity mesh for collision and shadowing and present solutions to integrate it with the color and LoD systems.
The resulting solution is fast enough to render frames stereoscopically for virtual reality systems.
Interactivity is naturally achieved through the collision mesh, which enables hand-interactions and inter-object collisions.
Overall, we claim the following contributions:
\begin{itemize}
    \item we present a model that enables photorealistic object appearance capture in natural environments using low-fidelity capture devices, such as mobile phones,
    \item it enables rendering at interactive framerates even in stereoscopic settings while supporting volumetric materials,
    \item seamless integration into a virtual scene of a game engine, including shadowing, collision and level-of-detail.
\end{itemize}
\section{Related Work}
\label{sec:related_work}

\paragraph{Object Capture.}
3D reconstructing objects and environments using photometric information has a long history~\cite{meydenbauer1867photometrographie,shape_from_image_streams_kanade1992}.
In recent years, the highest quality capture rig for this purpose is the light stage~\cite{light_stage}.
It uses many point lights mounted on a spherical hull to record an object under many lighting conditions with hard shadows, emphasizing the geometrical structure of the object and facilitating 3D reconstruction with a high level of detail.
However, it is costly to build and complex to operate and can not be used for in-the-wild captures.

With the release of the Kinect sensor, Microsoft made depth sensors readily available for consumers and researchers.
This inspired research on a variety of approaches that use depth information for 3D reconstruction~\cite{izadi2011kinectfusion,kinect_survey,kinect_dataset_2019,qi2016pointnet}.
However, with the recent popularity of radiance field approaches for 3D reconstruction, datasets recorded solely with consumer-grade mobile devices are starting to emerge~\cite{reizenstein21co3d}.
Photogrammetry has always tried to enable capture in in-the-wild scenarios: COLMAP~\cite{schoenberger2016sfm} and AliceVision~\cite{alicevision2021} are examples of free photogrammetry packages.
All of these packages provide camera calibration, triangulation and meshing tools, making them suitable to reconstruct textured object meshes from images or videos.
Still, pure mesh reconstructions lack view-dependent effects and cannot model transparent object parts.
Compared to radiance field representations, they often have to be manually `cleaned up' by artists because the reconstructed surfaces contain artifacts from a reconstruction pipeline that (1) can not model more complex effects violating the Lambertian assumption, such as specularities or subsurface scattering, and (2) is composed of multiple steps that are optimized separately and can not propagate information end-to-end.
This stands in contrast to radiance fields that capture viewpoint dependent effects and can be optimized end-to-end. 

\paragraph{Radiance field representations.} 

Neural Radiance Fields (NeRFs) model the plenoptic function as a coordinate-based function that is being aggregated using an emission-absorption model~\cite{mildenhall2020nerf}.
NeRFs make volume rendering computationally tractable in an end-to-end optimization using multiple images, yet using a model that can capture any viewpoint-dependent effects given enough model capacity.
Most existing works focus on modeling entire scenes and not objects and have problems modeling scenes with details close to the camera (foreground) as well as far away from the camera (background).
Two recent works~\cite{zhang2020nerf++,barron2022mip360} alleviate this problem by introducing explicit `background' modeling, yet still combine foreground and background in the same representation.
\cite{xie2021fignerf,kobayashi2022distilledfeaturefields,ranade2022ssdnerf} decompose foreground and background models to create a separate foreground model---a strategy that we apply as well to create high quality decompositions of objects vs. background.

Several works have improved upon the slow runtime of the original radiance field implementation: 
Plenoxels~\cite{yu2021plenoxels} and ReLU-Fields~\cite{karnewar2022relu} use a voxel grid encoding for better runtimes at training and test time and remove neural networks from the rendering process.
Instant-NGP~\cite{muller2022instant} uses a hash function instead, creating a volumetric storage for features that are processed by a small neural network; \cite{chen2022tensorf} uses volume factorization to reduce memory use.
We found a hybrid approach that lead to superior runtime as well as render performance: we use a regular grid for volumetric feature storage, and two MLPs to create density and viewpoint-dependent features for our encoding.
To achieve high quality results in real-world scenarios, the quality of the camera parameters is critical.
During 3D reconstruction gradients are readily available for every pixel, and they can be used to refine the camera parameters~\cite{lin2021barf,wang2021nerfminmin,jeong2021scnerf}.
Similar to these works, we propagate gradients from our 3D representation to the camera parameters.
For a comprehensive overview over recent work on radiance fields, we refer to~\cite{tewari2022advances}.
Overall, many existing radiance field approaches have too slow render speeds for interactive experiences or are not suitable for implementations using shading languages.

\paragraph{Real-time neural rendering.}
Several approaches have been created specifically to address rendering speed: PlenOctrees~\cite{yu2021plenoctrees} use an octree acceleration structure to speed up volume traversal at render time and store view-dependent information in a spherical harmonics encoding.
SNeRG~\cite{hedman2021snerg} uses a volumetric grid for feature storage and predicts a separation of diffuse and specular colors from those features.
Pulsar~\cite{lassner2021pulsar} is a sphere-based renderer with high render speeds that uses a slightly different ray accumulation function compared to NeRF, but can also be used for volumetric rendering of complex scenes.
Instead of using one large MLP to model radiance fields, they can be broken up into many small models to improve render speed~\cite{reiser2021kilonerf,rebain2021derf}.

Another approach to achieve speedups is to use fewer, but better-placed samples during volume sampling.
DONeRF~\cite{neff2021donerf} uses a learned sampling strategy (oracle network) for this purpose---however, this leads to problems at object boundaries and with fine details, a use case that we focus on.
Fast-NeRF~\cite{garbin2021fastnerf} uses another strategy: prediction caching.
This, however, is only applied at test time and has high memory requirements.
MobileNeRF~\cite{chen2022mobilenerf} distills trained radiance fields into a multi-plane mesh structure, which leads to very fast render results but sacrifices volumetric effects.
In comparison, our proposed method is faster than MobileNeRF (the fastest of the aforementioned approaches), shows relatively high speed for training as well as rendering and retains the ability to correctly render volumetric and viewpoint-dependent effects.

\section{Method}
\label{sec:method}

\begin{figure*}
    \centering
    \includegraphics[width=0.98\textwidth]{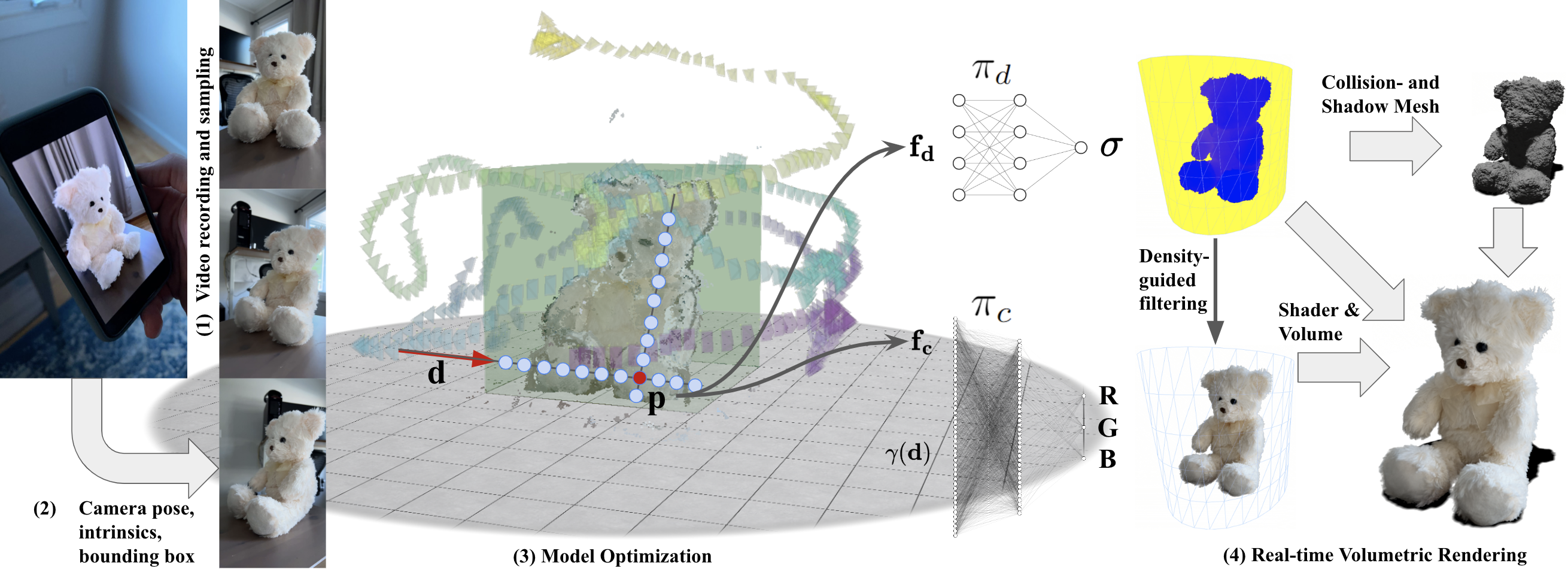} 
    \caption{\textbf{Method overview.} \textbf{(1)}~Objects can be casually recorded with a smartphone. Our captures usually take 90s or less. \textbf{(2)}~After an initial filtering and solve for camera pose, intrinsics and bounding box specification, \textbf{(3)}~an end-to-end optimization for 3D reconstruction is performed. For every sample point, we split the features into 4 density features that are processed using an MLP to produce $\sigma$, and 8 appearance features that are concatenated with a viewpoint encoding and processed by a separate MLP to produce color predictions. \textbf{(4)}~We export a shadow- and collision mesh and the feature volume and transpile the model to shader code for real-time volumetric rendering.
    }
    \label{fig:method_overview}
\end{figure*}

In the following sections, we describe our proposed capture- and rendering-processes, as well as the model architecture and engine integrations to take objects from the physical into a virtual world.

\subsection{Object Capture} \label{sec:capture}
The first step for the entire system is the object capture: we need to find a set of images suitable for reconstruction, camera intrinsics and extrinsics and an object boundary.
To keep this process as lightweight and efficient as possible, we design it to work with a low-fidelity camera, for example from a mobile phone, and require only a short video recording of the object (about 90s; we include an example capture in the suppl.~mat.).
We aim to support a wide variety of devices, hence we decided not to use any on-device camera tracking or vendor-specific reconstruction or SLAM frameworks.
We solely assume fixed exposure, white balance and sensitivity settings during object recording---these settings are trivial to set by the user using off-the-shelf recording applications.
The resulting video may still contain blurry frames that could hinder a detailed reconstruction.
To address this, we use a simple gradient-based motion blur detector that filters blurry frames and selects a subset of frames sampled at an interval of 6FPS.
These frames are processed by an off-the-shelf photogrammetry software that estimates approximate camera intrinsic parameters $\mathbf{K}$ (pre-calibration is possible as well) and poses $\mathbf{R}_k, \mathbf{t}_k$ for each frame $k$.
Using this initial rough solve and the resulting point cloud, we ask the user to provide a rough bounding box in 3D around the foreground object.
While it is possible to use an object segmentation method or a video salient object detector (for which we also present experiments in Sec.~\ref{sec:results}), we use this simple method for reliable, fine-grained user control.

\subsection{3D Reconstruction}

Given RGB frames with approximately estimated camera intrinsics $\mathbf{K}$ and poses $\mathbf{R}_k, \mathbf{t}_k$ for every frame $k$, we aim to optimize a volumetric representation to model the object appearance.
However, we are facing two problems: (1) we want to obtain the highest level of detail in the reconstructed object (the foreground), while still requiring a fairly detailed reconstruction of the background to create high quality segmentation and translucency effects; (2) the representation has to be separated after reconstruction to extract the foreground object.
To address that, we propose to use a two-layer volumetric representation similar to~\cite{xie2021fignerf,kobayashi2022distilledfeaturefields,ranade2022ssdnerf}.

\paragraph{Two-layer volumetric rendering.}
Overall, we represent the objects using a volumetric representation, following neural radiance fields~\cite{mildenhall2020nerf}.
To render a novel view, the color of each pixel is computed using volumetric integration along each pixel ray by using many sample points $\mathbf{p}_i = \mathbf{o} + \delta_i \mathbf{d}$, where $\mathbf{o}$ is the corresponding camera origin, $\mathbf{d}$ is the normalized pixel view direction, and $\delta_i$ is the current sample point's distance to the origin.
They can be computed for pixel $(x, y)$ in frame $k$ as:
\begin{align} \label{eq:o_and_d}
    \mathbf{o} &= \mathbf{t}_k, \\
    \mathbf{d} &= \mathbf{R}_k \mathbf{K}^{-1} (x, y, 1)^T.
\end{align}
The model consists of two modules: the density module $\mathcal{D}$ that takes point $\mathbf{p}_i$ as input and predicts its density value $\sigma_i = \mathcal{D} (\mathbf{p}_i)$, and the appearance module $\mathcal{C}$ that predicts view-dependent color $\mathbf{c}_i = \mathcal{C} (\mathbf{p}_i, \mathbf{d})$ given the point position $\mathbf{p}_i$ and view direction $\mathbf{d}$.
The pixel color is obtained by integrating these density and color values using numerical quadrature~\cite{max1995optical}:
\begin{equation} \label{eq:vol_integration}
    \mathbf{c} (\mathbf{o}, \mathbf{d}) = \sum_i w_i \mathbf{c}_i
\end{equation}
with sample weight $w_i$ computed as:
\begin{equation}
w_i = {\rm e}^{-\sum_{j < i} \sigma_j (t_{j+1} - t_j)} (1 - {\rm e}^{-\sigma_i (t_{i+1} - t_i)})
\end{equation}

This representation supports complex volumetric effects such as translucency and can, given a viewpoint-dependent model for $\mathbf{c}_i$, even capture subsurface scattering and reflections.
However, it couples the representation of the object with the environment.
Therefore, we split the model into two layers with separate density and appearance modules for the object $\mathcal{D}_o, \mathcal{C}_o$ and the environment $\mathcal{D}_e, \mathcal{C}_e$.

To compute the final color, we follow Eq.~\ref{eq:vol_integration}, and compute the density $\sigma_i$ and color $\mathbf{c}_i$ from object properties $\sigma_i^o$, $\mathbf{c}_i^o$ and environment properties $\sigma_i^e$, $\mathbf{c}_i^e$ as:
\begin{align}
    \sigma_i &= \sigma_i^o + \sigma_i^e, \\
    \mathbf{c}_i &= (\sigma_i^o \mathbf{c}_i^o + \sigma_i^e \mathbf{c}_i^e) / \max(\sigma_i, \epsilon),
\end{align}
where $\epsilon = 1e^{-10}$ is added solely for numerical stability. 
This formulation conveniently combines the contributions of object and environment layers directly in the aggregation.
If a 3D bounding box is available, any samples that fall within the box are sampled from the foreground layer model, and any samples that are outside are sampled from the background layer model.
Without bounding box we instead use predicted semantic masks to disambiguate the foreground from the background.
In that case, object and environment density estimates can be accumulated along the rays, assuming white radiance values, resulting in rendered object and environment masks.
These are enforced to match the predicted semantic masks, guiding the neural asset segmentation, similar to~\cite{ranade2022ssdnerf}.

To optimize the model, we take sensor pixel observations $\mathbf{\tilde{c}}$ and enforce that our model prediction from the rendering Eq.~\ref{eq:vol_integration} matches it by evaluating the $\ell_2$ color difference:
\begin{equation} \label{eq:rendering_loss}
    \mathcal{L} = \sum_{\mathbf{o}, \mathbf{d}} || \mathbf{\tilde{c}} (\mathbf{o}, \mathbf{d}) - \mathbf{c} (\mathbf{o}, \mathbf{d}) ||^2_2
\end{equation}

However, to make optimization robust and achieve the highest quality, we introduce additional improvements.

\paragraph{Camera pose and intrinsics refinement.}
The initial camera pose and intrinsics estimates can be inaccurate, and sub-pixel precision is needed for high-quality reconstruction. 
We refine the parameters for camera poses $\mathbf{R}_k, \mathbf{t}_k$ and pinhole intrinsics $\mathbf{K}$, and optimize them together with the object model. 
To ensure the rotation matrices $\mathbf{R}_k$ remain valid throughout the gradient-based optimization, we use a 6D parametrization~\cite{zhou2019continuity} for the rotations.

For a stable optimization we found it useful to include additional pose regularization with respect to the initial poses $\mathbf{\hat{R}}_k$ and $\mathbf{\hat{t}}_k$ penalizing strong deviations:
\begin{equation}
    \mathcal{L}_{\mathrm{pose}} = \sum_k ( \lambda_{\mathbf{R}} || \mathbf{R}_k - \mathbf{\hat{R}}_k ||^2_2 + \lambda_{\mathbf{t}} || \mathbf{t}_k - \mathbf{\hat{t}}_k ||^2_2 ).
\end{equation}
We use different learning rates for rotation, translation and the intrinsic matrix.
For further details on the training, learning rates and $\lambda_R$ and $\lambda_t$, please see the suppl.~mat.

\paragraph{Coarse-to-fine grid optimization.}
To improve model convergence and make the camera refinement more effective, we construct the feature grid in a coarse-to-fine manner, similar to~\cite{yu2021plenoxels}.
We start the model optimization with a grid resolution of $d = 32$, and upsample it trilinearly to a two times larger resolution every $5k$ iterations, up to the final dimension of $d = 256$.

\paragraph{Modelling unbounded environments.}
In everyday environments it is very common to make recordings in `unbounded' areas, i.e. where the maximum ranges for some rays are much larger than for the foreground object.
Failure to model these rays causes artefacts in the object representation and prevents complete background segmentation.
To extend our model to unbounded scenes, we add a 3D space contraction to the environment layer, which non-linearly warps all point coordinates along the ray from an unbounded interval $[1, \infty)$ to the bounded interval $[1, 2)$, as is detailed in~\cite{barron2022mip360}.

\paragraph{Weight regularization.}
Using only the photometric loss (Eq.~\ref{eq:rendering_loss}) for optimizing the radiance field can results in artefacts in `free space' when generating novel views.
Thus, we follow the weight regularization strategy from~\cite{barron2022mip360} encouraging every ray to either have zero weights along the entire ray, or to consolidate all weights into as small area as possible, resulting in an additional loss $\mathcal{L}_{\mathrm{weight}}.$

\paragraph{Adaptive point sampling.} 
For accurate appearance modelling a high number of sample points $p_i$ are required along each ray during optimization.
We use at most $N=1024$ samples; to distribute them effectively, we divide them into three regions: most samples are linearly allocated to the approximated foreground bounding box, with the remaining samples either in front of the object (linearly in depth) or behind the object (inversely proportional to depth).

\subsection{Real-time Rendering}
Rendering radiance fields in interactive environments requires super real-time performance: they are part of a larger rendering system, requiring time for scene logic, physics, \ldots, and potentially having to render twice for stereoscopic render targets in virtual or augmented reality.
In the original neural radiance field formulation~\cite{mildenhall2020nerf}, the density $\mathcal{D}$ and color $\mathcal{C}$ modules are modelled by single, large multi-layer perceptron (MLP). 
Evaluating this MLP for every point along the ray is prohibitively expensive, taking minutes per frame to render.
We tackle this problem in three steps: (1)~hardware-guided model design, (2)~density-guided sample filtering and (3)~model transpilation into optimized shader code.

\paragraph{Hardware-guided model design.}
\begin{figure}
    \centering
    \includegraphics[width=0.98\columnwidth]{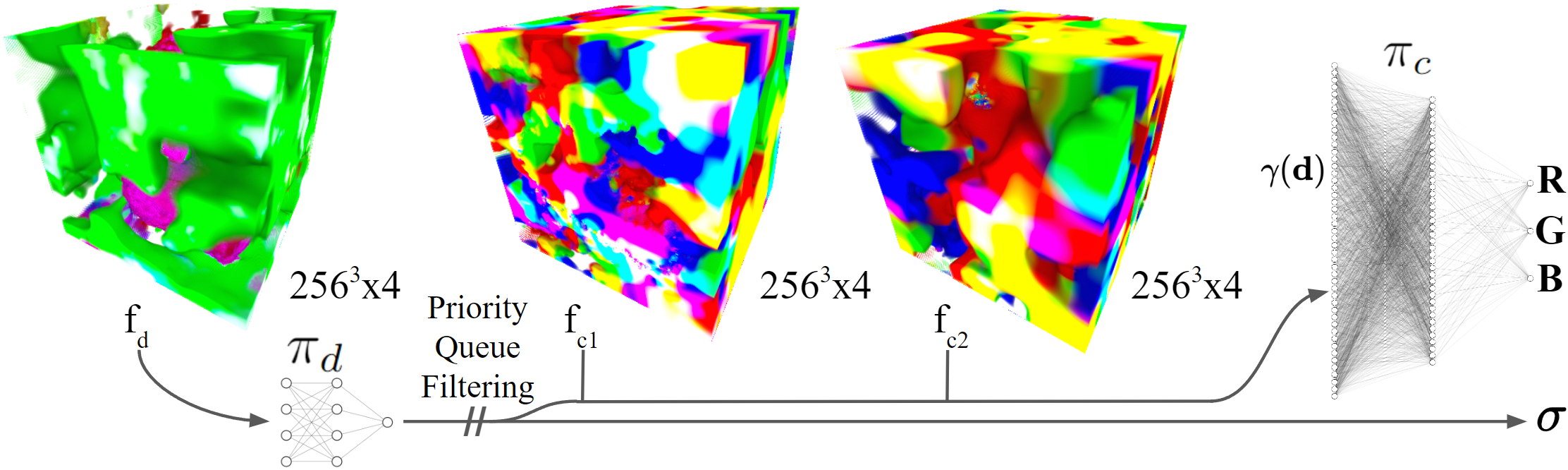}
    \vspace*{-0.2cm}
    \caption{\textbf{Model execution graph.} All texture reads are hardware-supported as a single operation including trilinear interpolation. Reading $f_d$ is used to evaluate $\pi_d$, potentially for several/all points to do priority-queue based filtering. Only for `valuable' points, texture reads for $f_{c1}$ and $f_{c2}$ are performed and $\pi_c$ executed.}
    \label{fig:actiongraph}
\end{figure}
Current generation GPUs provide several gigabytes of memory and hardware-support for texture readouts, including interpolation of voxel values.
This is an incredibly powerful and fast way to store large amounts of volumetric data and the inspiration for recent neural network free radiance field models such as ReLU fields~\cite{karnewar2022relu}.
ReLU fields store features in a volume structure, trilinearly interpolate them for sample point readouts and solely use a rectified linear unit (ReLU) as non-linearity to produce density or viewpoint-dependent effects.
This leads to very fast processing times, but we found the visual quality insufficient in many scenarios: details were not always reproduced faithfully, or the grid resolution had to be increased which leads to very high memory requirements due to the cubic growth of storage requirements (multiplied with the number of feature channels per point).
However, we found a combination of a 3D feature grid $\psi$ of dimension $f \times d^3$ and very small MLP decoders $\pi_d$ and $\pi_c$ (for density and color, respectively) to be even more efficient, while at the same time obtaining higher quality novel views.
We assume the higher computational efficiency is due to the more compact representation in memory which leads to better cache coherence for sampling.
The MLPs allow us to reduce the grid size to $256^3$ and set the number of features as low as 4 for the density and 8 for the appearance predictor, split into $\mathbf{f_d}\in\mathbb{R}^4$ and $\mathbf{f_c}\in\mathbb{R}^8$.
It is critical to use multiples of four for the number of features for each network to maximize texture efficiency: one hardware-accelerated texture read fetches four channel values.
Hence, for each sample point $\mathbf{p}_i$, its features $\mathbf{f}$ are obtained from the trilinearly interpolated feature grid $\psi$.
The sample's density and color values are obtained by processing the features by MLPs:
\begin{align}
    \sigma_i &= \pi_d (\mathbf{f_d}), \\
    \mathbf{c}_i &= \pi_c (\mathbf{f_c}, \gamma(\mathbf{d})).
\end{align}
Similar to~\cite{mildenhall2020nerf}, we use a \emph{positional encoding} function $\gamma(.)$ for the view direction $\mathbf{d}$, with 6 frequencies.

\paragraph{Density-guided sample filtering.}
A major contribution to model runtime is the number of evaluated feature points: when determining the runtime of the rendering system, the number of pixels/rays is multiplied by the number of samples along each ray.
Hence, we aim to reduce the number of sample point evaluations.
We considered several ways of sample culling, but identified most of them to either fail for rays close to object boundaries, for rays with multiple major intersections or found them to be not suitable for efficient execution on a GPU because of excessive branching requirements.
We found density-guided sample filtering to improve the render performance a lot without sacrificing quality and being a good fit for the GPU execution model.
The decision to split the non-viewpoint dependent density calculation in feature and model space is the first critical part: it makes obtaining $\sigma_i$ for a sample point very cheap.
We model $\pi_d$ as a single $4\times4$ fully-connected layer, followed by a SoftPlus activation~\cite{barron2022mip360}.
This means it can be evaluated by a single texture fetch operation followed by five SIMD dot products and the SoftPlus non-linearity---we can easily afford executing this operation along the entire ray, or until we meet an aggregation density target.
On the other hand, $\pi_c$, the color decoder, must be larger and more expensive to model complex viewpoint dependent appearance and is more expensive to evaluate. 
Hence, we incrementally build up a priority queue of weight values $w_i$ along the ray using $\pi_d$ densely, and evaluate the appearance only at the samples with the highest blending weights.

\paragraph{Transpilation into efficient shader code.}
Integrating radiance fields seamlessly into current render and game engines requires optimized code, integrated with shading language standard that is portable to many platforms.
We decide to support HLSL and GLSL, making a wide variety of render engines including web browsers (thanks to WebGL supporting GLSL) to target platforms.
We implemented a transpiler that takes PyTorch modules and generates performant shader code for a pixel shader in HLSL or GLSL.
Within the rendering engine, we can create a proxy volume and assign the pixel shader to it, automatically running it for every pixel that lies on the respective volume and rendering the reconstructed object.
The transpiler runs several optimizations: it optimizes the use of registers to minimize the amount of register memory required, and it groups operations into vectorized SIMD operations to make use of the wide GPU instruction set.

\begin{figure*}
    \centering
    \includegraphics[width=0.98\textwidth]{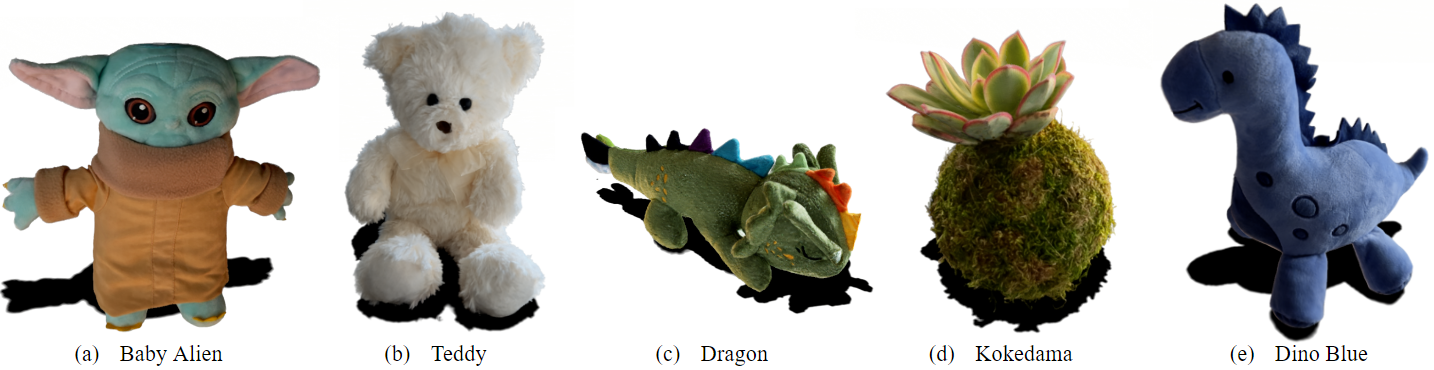}
    \vspace*{-0.2cm}
    \caption{\textbf{Test views of reconstructed benchmark objects.} All objects have been reconstructed from mobile phone recordings and have been picked as a set of hard test objects. (a) has lots of wrinkles and concave areas, (b) transparency and fur, (c) viewpoint dependent effects (sparkles on surface), (d) intricate surface details, (e) viewpoint dependent shading effects.}
    \label{fig:benchmark_objects}
\end{figure*}

\subsection{Integration into a Game Engine}
\label{sec:integration_game_engine}
Achieving interactivity requires additional integrations with the game engine.

\paragraph{Level-of-detail.}
When we interact with volumetric objects in a game engine, especially in virtual reality, where the distances to the object can vary a lot, the aliasing artefacts can be quite severe.
Furthermore, even though the neural asset rendering is very efficient, querying high-resolution feature grids can become expensive with a lot of objects that are visible at the same time in the scene (\eg, when going further away from the objects). %
To address both of these problems, we introduce a \emph{hierarchy of feature grids} by using the full-resolution 3D feature grid and trilinearly downsampling it to multiple levels of lower resolutions (similar to mipmaps for textures in graphics).
When querying features, we adaptively pick the two grid levels appropriate depending on the distance to the viewpoint, and linearly interpolate their feature values.
This makes the feature querying for far-away objects faster and also eliminates aliasing issues thanks to the implicit lowpass filtering.

\paragraph{Collisions and shadow mesh.}
We aim for a fully seamless experience where objects can be interacted with and can interact with each other (for instance by colliding).
This is critical for experiences in which a physics engine dictates object placement in the world, or to enable natural object interactions through hand tracking in augmented and virtual reality.
To achieve that, we additionally compute a coarse collision mesh for every object and register it with the object volume.
This mesh can be be directly extracted from the objects density field using marching cubes~\cite{lorensen1987marching}.
We use the same mesh as a shadow mesh to cast dynamic shadows to the environment.

\paragraph{Colors and blending.}
Modern game engines use a wide gamut linear color space for compositing and map it only during a final render pass to the respective device color space.
This is important to take into account to achieve correct results for radiance field rendering: after aggregation, we map the predicted result from sRGB to a linear color space, normalize for scene exposure and let the game engine map it to the output color space.
Furthermore, we disable wide gamut mapping, though we note that it is possible to use it in principle as long as it is being accounted for in the render process.
To enable correct alpha-blending of the object with the rest of the scene, the alpha values have to be correctly treated as pre-multiplied alpha.

\section{Results}
\label{sec:results}

In this section we extensively evaluate the components of our approach, and compare our real-time rendering runtime and quality with other state-of-the-art approaches.

\paragraph{Experimental setting.}
We evaluate our approach on several datasets on the task of novel view synthesis, with a fixed split of train/val/test views. 
For each test view, we compare the rendered pixel colors with the raw observations, and evaluate peak signal-to-noise ratio (PSNR) and structural similarity index measure (SSIM).
We report the average metrics over all test views.

\begin{table}
  \centering
  \begin{tabular}{lcc}
    \toprule
    \textbf{Approach} & \textbf{PSNR} & \textbf{SSIM} \\
    \midrule
    Single-layered representation & $31.77$ & $0.9414$ \\
    W/o coarse-to-fine grid optimization & $25.70$ &  $0.8704$\\
    W/o camera refinement & \cellcolor{bronze!25}$33.39$ & \cellcolor{silver!25}$0.9584$ \\
    W/o weight regularization & $33.04$ & $0.9578$ \\
    W/o adaptive sampling & \cellcolor{silver!25}$33.94$ & \cellcolor{bronze!25}$0.9581$ \\
    \midrule
    Ours & \cellcolor{gold!25}$34.25$ & \cellcolor{gold!25}$0.9614$ \\
    \bottomrule
  \end{tabular}
  \vspace*{-0.2cm}
  \caption{
  \textbf{Ablation study} of our approach averaged over five benchmark object reconstructions.
  }
  \label{tab:real_ablations}
  \vspace*{-0.1cm}
\end{table}

\begin{table}
  \centering
  \begin{tabular}{lccccc}
    \toprule
    \textbf{Approach} & \textbf{PSNR} & \textbf{FPS} & \textbf{CR} & \textbf{VE} \\
    \midrule
    ReLU-Fields~\cite{karnewar2022relu} & $30.04$ & $10.05$ & & \\
    KiloNeRF~\cite{reiser2021kilonerf} & \cellcolor{bronze!25}$31.00$ & $38.46$ & & \checkmark  \\
    FastNeRF~\cite{garbin2021fastnerf} & $29.97$ & \cellcolor{bronze!25}$238.10$ & & \\
    PlenOctrees~\cite{yu2021plenoctrees} & \cellcolor{gold!25}$31.71$ & $167.68$ &  & \\
    SNeRG~\cite{hedman2021snerg} & $30.38$ & $207.26$ & \ & \checkmark \\
    MobileNeRF~\cite{chen2022mobilenerf} & $30.99$ & \cellcolor{silver!25}$744.91$ &  &  \\
    \midrule
    Ours & \cellcolor{silver!25}$31.11$ & \cellcolor{gold!25}$813.01$ & \checkmark & \checkmark \\
    \bottomrule
  \end{tabular}
 \vspace*{-0.2cm}
  \caption{
  \textbf{Quantitative comparison of different interactive rendering methods on the NeRF-synthetic~\cite{mildenhall2020nerf} dataset}. We additionally note down which approaches support camera refinement (CR) and showcase perceived volumetric and high-frequency viewpoint dependent effects (VE), both very important for reconstructing real-world objects using phone recordings.
  }
  \vspace*{-0.1cm}
  \label{tab:synthetic_comparison}
\end{table}

\paragraph{Ablation of model components.}
We trained our approach on five recordings of real-world objects with particularly challenging materials in everyday environments, resulting in neural assets featuring fur, translucent parts and thin structures.
The reconstructions using our method are visualized in Fig.~\ref{fig:benchmark_objects}.
In Tab.~\ref{tab:real_ablations} we evaluate different model components, showing that each of them is important to achieve high quality reconstruction.
We additionally provide a qualitative preview of interactions with these neural assets in VR, using a headset and hand tracking in the suppl.~mat.
The reconstruction for all of these objects is automatic and we use the same hyperparameters for all of them, demonstrating the generality of the approach.

\paragraph{Comparison with baseline methods.}
In Tab.~\ref{tab:synthetic_comparison} we compare the rendering of our neural assets with existing neural rendering methods that support (close to) real-time rendering performance, on the test set of the Synthetic-NeRF dataset~\cite{mildenhall2020nerf}. 
Our approach achieves the best runtime performance, while still attaining a very high reconstruction quality.
The runtime was measured by rendering novel views at resolution of $800 \times 800$, using an RTX3080 GPU.

\begin{figure}
    \centering
    \includegraphics[width=\linewidth]{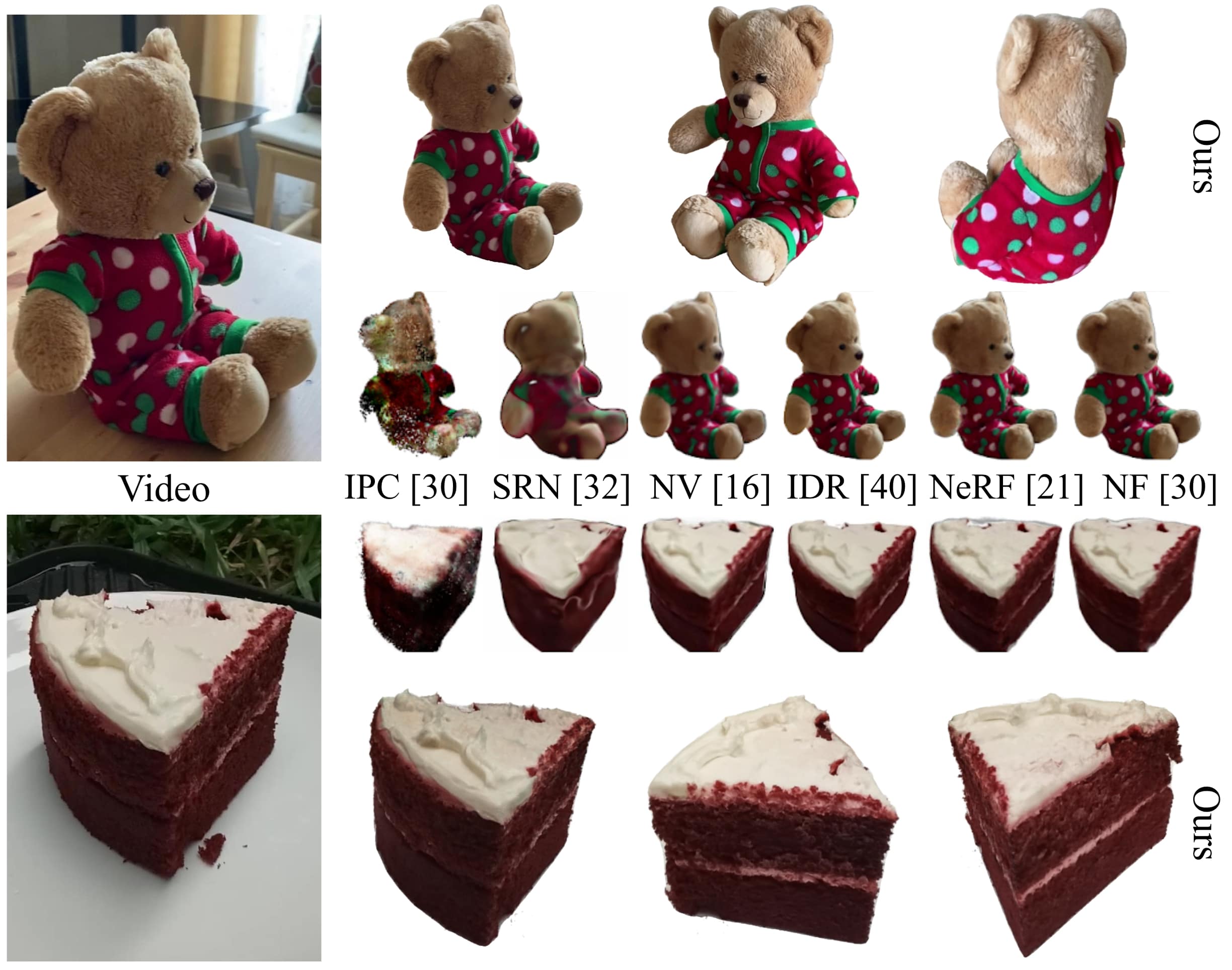} 
    \caption{\textbf{Qualitative comparisons} of our approach with the reconstruction methods~\cite{reizenstein21co3d, sitzmann2019scene, lombardi2019neural, yariv2020multiview, mildenhall2020nerf} on sequences from the CO3D dataset~\cite{reizenstein21co3d}.}
    \label{fig:co3d_results}
\end{figure}

\paragraph{Using semantic mask predictions.}
If an estimated object bounding box is not available, \eg, in the lightweight captures of the publicly available CO3D dataset~\cite{reizenstein21co3d}, we instead rely on semantic object masks, which can be predicted by an off-the-shelf foreground segmentation network or video salient object detection model~\cite{xu2021locate}.
In Fig.~\ref{fig:co3d_results} we show novel views on sequences from the CO3D dataset and compare them with other neural rendering approaches.

\paragraph{Rendering efficiency.}

\begin{table}
  \centering
  \resizebox{0.98\columnwidth}{!}{
  \begin{tabular}{lccc}
    \toprule
    \textbf{Renderer} & \textbf{Hardware} & \textbf{Resolution}  & \textbf{FPS} \\
    \midrule
    PyTorch & V100 & $1920 \times 1080$ & $0.53$ \\
    Unreal Shader & RTX3080 & $1920 \times 1080$ & $423.73$ \\
    Unreal Frame & RTX3080 & $1920 \times 1080$ & $140.06$ \\
    WebGL Shader & RTX3080 & $3840 \times 2160$ & $59.40^\dagger$  \\
    WebGL Shader & M1 Pro & $3456 \times 2234$ & $62.40$ \\
    Quest Pro Shader & RTX3080 & $3840 \times 1800$ &  $150.60$ \\
    Quest Pro Frame & RTX3080 & $3840 \times 1800$ &  $95.15$ \\
    \bottomrule
  \end{tabular}
  }
  \vspace*{-0.3cm}
  \caption{
  \textbf{Rendering runtime} of our automatically generated shader code compared to PyTorch, and evaluated on different hardware and resolutions. $^\dagger$WebGL was capped to 60 FPS on this machine.
  }
  \label{tab:timings}
\end{table}

We generated efficient shader code for different environments, and extensively evaluated the rendering efficiency using a variety of hardware.
The timings are summarized in Tab.~\ref{tab:timings}. 
When the assets are rendered in a game engine, then beside the shader code that executes the volumetric rendering, we also report the frame time, which additionally includes the dynamic shadow visualization, collision resolution, etc.
It's interesting that at certain resolutions these classical mesh-based operations can take a lot of time compared to volumetric rendering.

\paragraph{Limitations.}
While our neural assets support photorealistic appearance, the feature grids that enable efficient rendering of detailed effects use more GPU memory than mesh objects (\eg, at the highest resolution, the grids need about 800 MB). 
An interesting future direction would be compressing the features further, \eg by exploring dense grid decomposition into 2D and 1D representations~\cite{chen2022tensorf}.
Furthermore, while our approach supports natural interactions with the neural objects, the objects are moving only rigidly.
It would be interesting to extend our method to a deformable NeRF representation~\cite{tretschk2021non, park2021nerfies, park2021hypernerf}, and add realistic deformations during interactions.
Another important addition for our models is editability: by storing features on the voxel grid, it is easy to enable local editing in principle.
However, a mapping betweeen features and intended colors or material properties is needed, and there exist many possibilities considering the latest advances in generative radiance fields~\cite{dreamfusion}.

\section{Conclusion}
\label{sec:conclusion}
In this paper, we introduced Neural Assets: a pipeline and model to create neural radiance field representations of objects and use them in interactive environments.
We created a new model architecture, suitable to model high quality viewpoint dependent and transparency effects that can be rendered in super real-time (greater than 200 FPS on consumer grade GPUs).
To achieve this speed, we propose to use a transpiler and convert the model directly to shader code and textures---components that are readily available in production-ready game engines.
On this end, we present solutions for full integration with interactive systems: shadows, physics and level-of-detail, tackling many practical issues and making the use of radiance fields in environments with mesh representations seamless and attractive.
The combination of visual quality of the real-world reconstructions and the efficiency gains of the presented model are outperforming the state-of-the art and translate to many existing 3D reconstruction pipelines.
Its rendering speed combined with the visual quality of the volumetrically rendered radiance fields make it an attractive model for many applications.

\iftoggle{cvprfinal}{%
\paragraph{Acknowledgements.} We thank D. Roble for his great support of the project; O.~Maury, C.~Marshall and M.~David for the inspiring discussions and feedback, Z.~Dong, Z.~Li and team for the excellent collaboration towards our demos; C. McCard and A. Ferguson for the design of the demo scenes.
}{}

\iftoggle{cvprfinal}{%
\begin{appendices}

\section{Optimization Details}

In this section we provide further details about model optimization.

\paragraph{Hyperparameters for training.}
We use an ADAM optimizer with momentum of $0.9$ and $\beta_1 = 0.82, \beta_2 = 0.97$.
The batch size is set to $8192$ rays, and we sample $N = 1024$ samples along each ray.
The model is trained for a total of $500$k iterations.
We set different learning rates for different parts of the model: $0.0024$ for the decoder, $0.027$ for the feature grid, \num{7e-6} for rotations, \num{3e-5} for translations and \num{2e-6} for pinhole intrinsics.
We apply an exponential decay of $0.035$ over $350$k iterations to learning rate.
To balance the loss terms, we use $\lambda_\mathbf{R} = 0.00046$, $\lambda_\mathbf{t} = 0.0018$ and $\lambda_\mathrm{weight} = 5\times 10^{-4}$.

\paragraph{Adaptive sampling.}
In the method section of the main paper, we introduce \emph{adaptive point sampling}, which makes a better use of limited $N=1024$ point samples along each ray, by focusing most samples in the approximate object bounding box. 
A visualization of the sampling is provided in Fig.~\ref{fig:sampling}. 
We distribute the samples in the following way:
\begin{itemize}
    \item $7$\% of samples in front of the object, using linear depth sampling (\emph{red});
    \item $63$\% of samples in the object bounding box, using linear depth sampling (\emph{green});
    \item $30$\% of samples in the range further away from the object, using inverse depth sampling (\emph{blue}).
\end{itemize}

\begin{figure}  
    \centering
    \includegraphics[width=\linewidth]{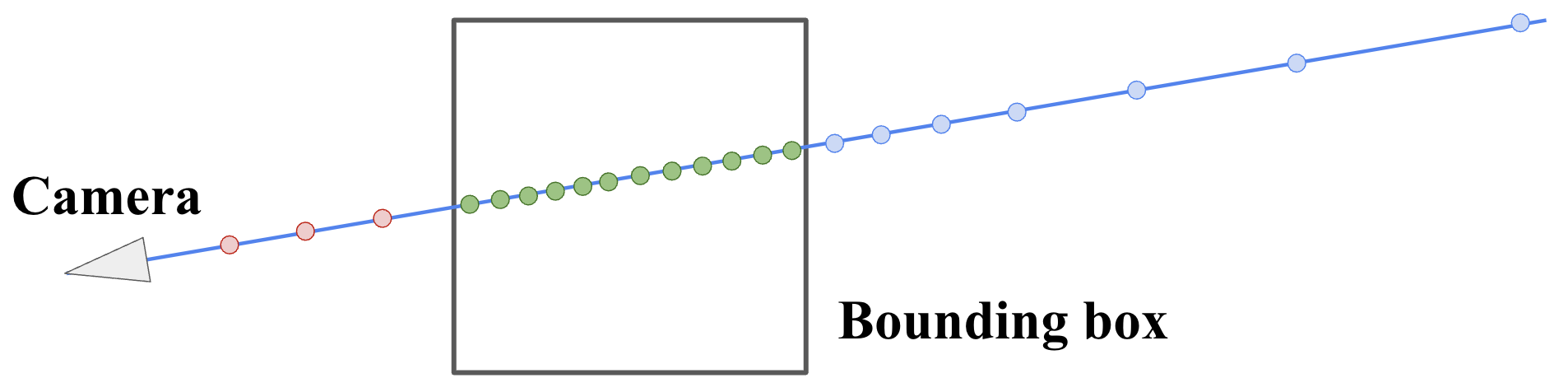} 
    \caption{\textbf{Illustration of adaptive point sampling.} We differentiate samples in front of the object (\emph{red}), inside the bounding box (\emph{green}) and further away from the object (\emph{blue}).}
    \label{fig:sampling}
\end{figure}

\section{Architecture Variations}
While we kept most architecture details consistent across different data sources, we did find it useful to differentiate the appearance decoder size for Synthetic-NeRF~\cite{mildenhall2020nerf} and real-world recordings.
Synthetic sequences include some scenes with multiple objects (e.g. \emph{Materials} sequence), which benefit from a larger appearance decoder (using a linear layer with width of $40$) but perform well with fewer samples per ray ($5$).
This delivers high-quality novel views even with the low number of samples (achieving the reported PSNR of $31.11$).
On the other hand, we found that real-world sequences benefit from a larger sample count to model translucency within objects and high-quality blending around their boundaries.
Hence, we use a smaller decoder (with a linear layer width of $12$) and 20 samples along each ray with equivalent runtime performance; a model with this architecture achieves 30.80 PSNR on the Synthetic-NeRF scenes.

\section{Visualizations}
We visualize different examples of hand interactions with virtual assets in Fig.~\ref{fig:interaction_visualization}.

\begin{figure}
    \centering
    \includegraphics[width=\linewidth]{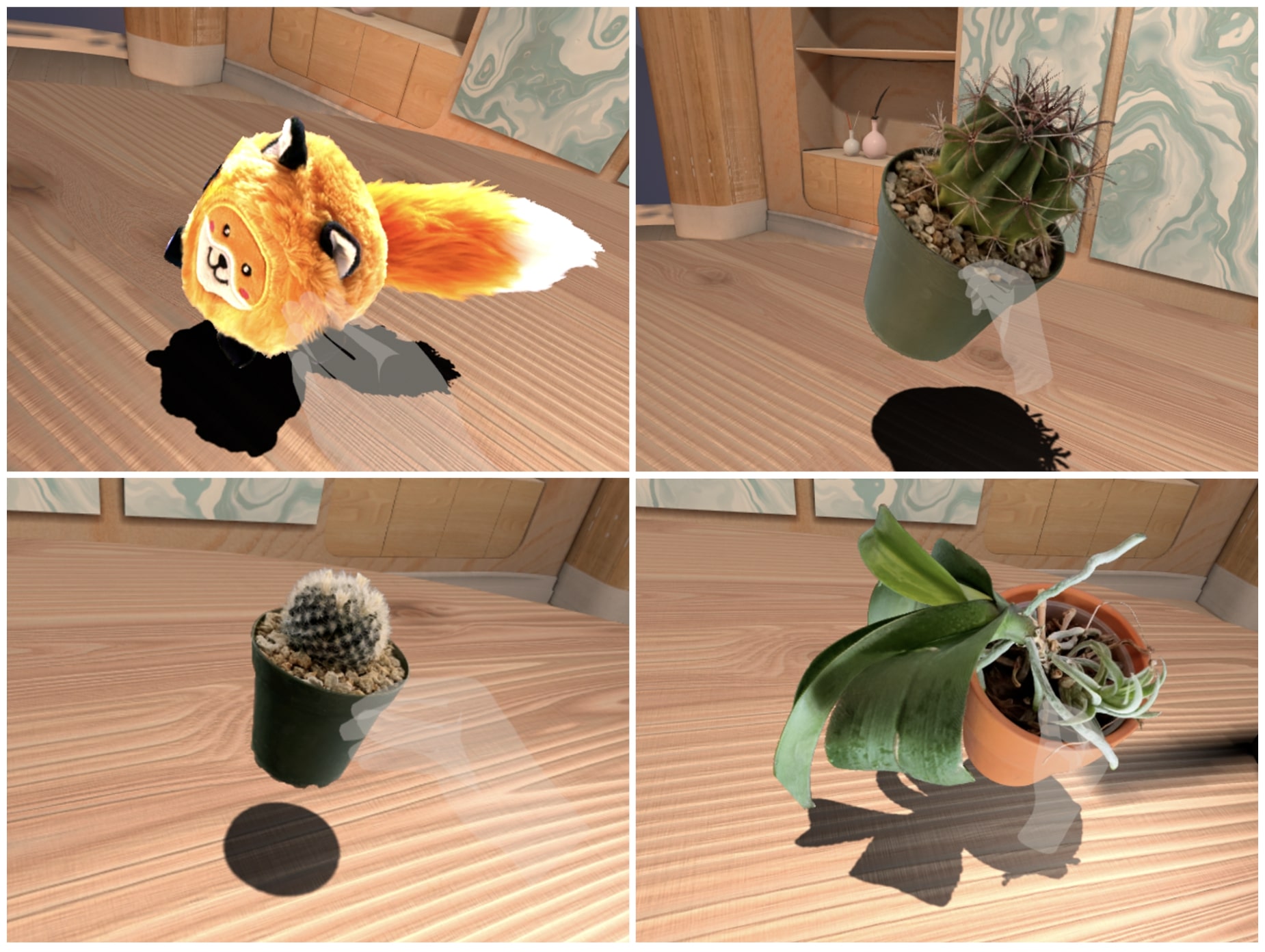} 
    \caption{Visualizations of hand interactions with neural assets in a virtual environment.}
    \label{fig:interaction_visualization}
\end{figure}
\end{appendices}
}

{\small
\bibliographystyle{ieee_fullname}
\bibliography{egbib}
}

\end{document}